\documentclass{article}

\usepackage{microtype}
\usepackage{graphicx}
\usepackage{comment}
\usepackage{amsmath}
\usepackage{subfigure}
\usepackage[square,sort,comma]{natbib}
\usepackage{booktabs} 
\usepackage[utf8]{inputenc} 
\usepackage[T1]{fontenc}    
\usepackage[hidelinks
]{hyperref}       
\usepackage{url}            
\usepackage{amsfonts}       
\usepackage{nicefrac}       
\usepackage{microtype}      
\usepackage{lipsum}
\usepackage{hyperref}

\newcount\JN
\ifnum\JN=1
    \usepackage{icml2023}
    \icmltitlerunning{The DeepCAR Method: Forecasting Time-Series Data That Have Change Points}
\fi
\ifnum\JN=1
    \newcommand{\figurewidth}{\columnwidth}
\fi
\ifnum\JN=2
    \newcommand{\figurewidth}{0.7\columnwidth}
\fi

\ifnum\JN=2
    \usepackage{algorithm}
    \usepackage{algorithmic}
\fi

\ifnum\JN=1
\twocolumn[
\icmltitle{The DeepCAR Method: Forecasting Time-Series Data That Have Change Points
}

\newcommand{\ourmethod}{BatchCP}
\begin{icmlauthorlist}
\icmlauthor{Ayla Jungbluth}{to}
\icmlauthor{Johannes Lederer}{to}

\end{icmlauthorlist}

\icmlaffiliation{to}{Department of Mathematics, Ruhr-University Bochum, Germany}

\icmlcorrespondingauthor{Ayla Jungbluth}{ayla.jungbluth@rub.de}
\icmlcorrespondingauthor{Johannes Lederer}{johannes.lederer@rub.de}

\icmlkeywords{Machine Learning, ICML}

]

\printAffiliationsAndNotice{ } 

\else

\newcommand{\ourmethod}{BatchCP}
\usepackage{arxiv}
\title{The DeepCAR Method: Forecasting Time-Series Data That Have Change Points
}

\author{
 Ayla Jungbluth \\
  Department of Mathematics\\
  Ruhr-University Bochum\\
  Bochum 44801, Germany \\
  \texttt{ayla.jungbluth@rub.de} \\
   \And
 Johannes Lederer \\
  Department of Mathematics\\
  Ruhr-University Bochum\\
  Bochum 44801, Germany  \\
  \texttt{johannes.lederer@rub.de} \\
  }
  
\begin{document}

\maketitle
\fi
\begin{abstract}
Many methods for time-series forecasting are known in classical statistics, 
such as autoregression, moving averages, and exponential smoothing.
The DeepAR framework is a novel, recent approach for time-series forecasting based on deep learning.
DeepAR has shown very promising results already.
However, time series often have change points,
which can degrade the DeepAR's prediction performance substantially.
This paper extends the DeepAR framework by detecting and including those change points.
We show that our method performs as well as standard DeepAR when there are no change points and considerably better when there are change points.
More generally, 
we show that the batch size provides an effective and surprisingly simple way to deal with change points in DeepAR, Transformers, and other modern forecasting models.


\end{abstract}

\newcommand{\jl}[1]{\textcolor{red}{#1}}
\section{Introduction}
\subsection{The DeepAR Forecasting Algorithm}
Time-series forecasting has become textbook material \citep{ts,deep2},
including classical approaches such as moving  averages and exponential smoothing \citep{book1}.
A much newer development is the use of recurrent neural networks \citep{ts44,art9}.
An architecture particularly suited for this are LSTMs \cite{book2}.
A prominent example of time-series forecasting with LSTM-based neural networks is the DeepAR algorithm \cite{deep,glu}.


One of the main features of DeepAR is that it can combine related time series---rather than considering each time series by itself.
The model produces probabilistic forecasts based on Gaussian (or negative binomial) likelihoods \citep{sny,matt,art10}. 
A key aspect of the DeepAR framework is the selection of the batches in the training process. 
Multiple training instances are created for each time series by selecting windows with different starting points from the original time series. 
It is ensured that the entire prediction range is covered when choosing these windows;
in particular,
the starting point could lie before the beginning of the time series,
in which case the unobserved targets are filled with zeros.

At each time step, 
the next step is to be predicted. 
The network receives the previous observations together with a set of covariates as input. The information is passed through the hidden layer to the likelihood function. 
The error is calculated during training using the current parameterization of the likelihood function. Thus, when performing backpropagation, the weights are updated and the values optimized.

After the weights of the network are trained, the forward propagation is performed using the previous input to obtain the distribution parameters $\mu$ (mean) and $\sigma$ (standard deviation) of the Gaussian likelihood.

\subsection{Limitation: Change Points}


Time-series data often have change points.
Frequently,
these change points are the result of manual interventions,
such as resetting a machine after maintenance
or replacing players or coaches in a football team.
But sometimes, these change points can also have more systematic reasons,
such as the end of a football season.
The simplest way to deal with these change points is to ignore the time periods around them,
but this approach potentially dismisses large amounts of relevant data.
Our question is, therefore, if we can account for  change points without losing predictive quality---or even improving prediction.

An example of a time series of machine data with change points can be seen in Figure~\ref{cpvv}. 
It represents the hydraulic oil pollution of a shredder machine. 
The points, where the time series drops from higher to lower values indicate replacements of the machine's pollution filter.
The change points themselves do not have a pattern and, therefore, are not necessarily predictable.
%
\begin{figure}[ht]
\vskip 0.2in
\begin{center}
\centerline{\includegraphics[width=\figurewidth]{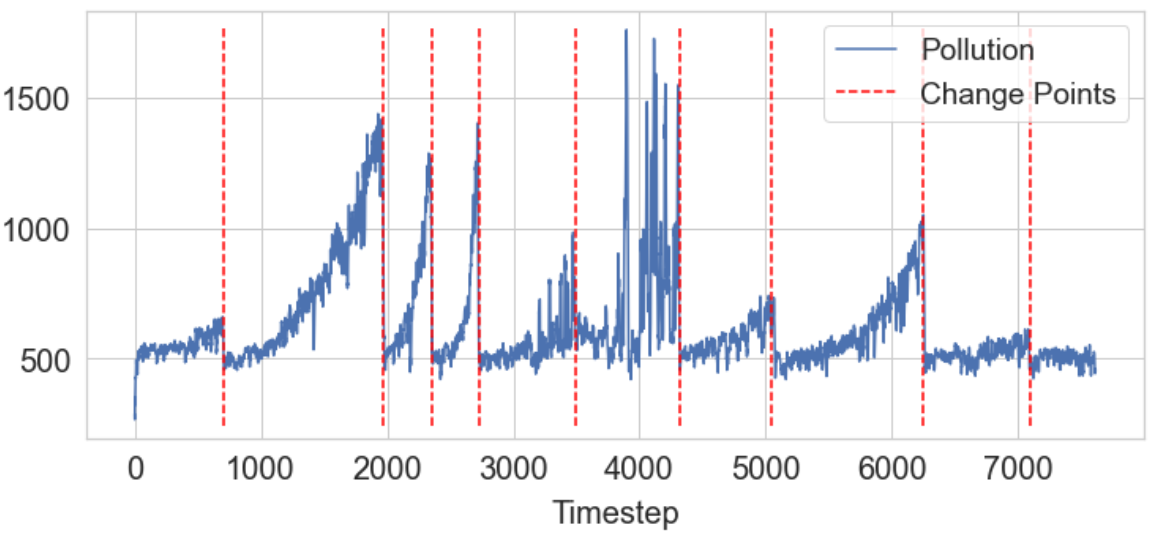}}
\caption{Example of the pollution time series (blue) with manual change points (red). The change points indicate machine maintenance. The behavior of the pollution data shows that it was replaced at times with change points. The change points are not at regular intervals and do not show seasonal behavior since the machine is not used consistently and has irregular deterioration. The change points are located at the time indices $700, 1970, 2350, 2730, 3500, 4320, 5050, 6250, 7100$.}
\label{cpvv}
\end{center}
\vskip -0.2in
\end{figure}
The vanilla DeepAR model would try to learn the change points in the time series:
indeed,
the algorithm takes batches from the time series, no matter if there is a change point in it or not.
This can deteriorate the forecasts and predict change points that do not exist.

To conclude,
DeepAR requires little manual feature engineering and recognizes seasonality,
but it does not incorporate change points correctly.
More generally,
we are not aware of any deep-learning-based method for time-series forecasting that integrates change points.



\label{submission}
\subsection{Research Question and Solution}
While, as discussed earlier,
change points are common in time-series data,
there is no accepted method for handling them in forecasting.
Our goal is, therefore, to interweave change-point detection with DeepAR and other forecasting methods.
By focusing on the training batches,
we will be able to do so while (i)~preserving the original forecasting methods' characteristics and (ii)~preserving the implementations.

Besides those two features, a strength of our approach is that  (iii)~it allows for the inclusion of any method for change-point detection.
This provides access to a vast pool of existing methods and potential future methods.

\section{Methods}
Since we refer to the DeepAR algorithm for our method, we will introduce it briefly.
\subsection{Mathematical Description of the DeepAR Model}
The following presents the autoregressive recurrent network architecture of the DeepAR model and the training process \cite{deep}.
\subsubsection{Model}
The value of time series $i$ at time $t$ is denoted by $z_{i,t}$. 
With given past data
\begin{equation*}
    [z_{i,1}, . . . , z_{i,t_0-2}, z_{i,t_0-1}] := \textbf{z}_{i,1:t_0-1},
\end{equation*}
the conditional distribution
\begin{equation*}
    P(\textbf{z}_{i,t_{0}:T} \vert \textbf{z}_{i,1:t_{0-1}}, \textbf{x}_{i,1:T} )
\end{equation*}
of the future of each time series 
\begin{equation*}
    [z_{i,t_0}, z_{i,t_0+1}, . . . , z_{i,T} ] := \textbf{z}_{i,t_0:T},
\end{equation*}is to be modeled.
Here $t_0$ denotes  the  time  point  from  which we assume that $z_{i,t}$ is unknown at prediction time. The covariates $\textbf{x}_{i,1:T}$ are  assumed  to  be known for all time points.
For the past, we write the time ranges $[1, t_0 - 1]$, called the conditioning range. For the future, we write $[t_0, T]$ as prediction range, respectively.
We assume that the model distribution 
\begin{equation*}
    Q_\Theta (\textbf{z}_{i,t_0:T} \vert \textbf{z}_{i,1:t_0-1}, \textbf{x}_{i,1:T} )
\end{equation*} consists of a product of likelihood factors 

\begin{multline*}
    Q_\Theta(\textbf{z}_{i,t_0:T} \vert \textbf{z}_{i,1:t_0-1}, \textbf{x}_{i,1:T} )
    = \prod^T_{t=t_0}
Q_\Theta (z_{i,t}\vert \textbf{z}_{i,1:t-1}, \textbf{x}_{i,1:T})
    = \prod^T_{t=t_0}\ell(z_{i,t} \vert \theta(\textbf{h}_{i,t}, \Theta))\,,
\end{multline*}
parameterized by the output $\textbf{h}_{i,t}$ of an autoregressive recurrent network 
\begin{equation*}
    \operatorname{\textbf{h}}_{i,t} = h (\operatorname{\textbf{h}}_{i,t-1}, z_{i,t-1}, \textbf{x}_{i,t}, \Theta )\,.
\end{equation*}
Here $h$ denotes a function implemented by a multilayer recurrent neural network with LSTM cells. We fed the network output $\textbf{h}_{i,t}$ into a function $\theta (\textbf{h}_{i,t}, \Theta)$, which then builds the parameters of the fixed distribution $\ell(z_{i,t}\vert \theta(\textbf{h}_{i,t}))$.


\subsubsection{Training}
A time-series data set $\{\textbf{z}_{i,1:T} \}_{i=1,...,N}$ and associated covariates $\textbf{x}_{i,1:T}$ are obtained by choosing a time range such that $z_{i,t}$ is known in the prediction range. The parameters $\Theta$ of the model consist of the parameters of the RNN $h(\cdot)$ and the parameters of $\theta(\cdot)$. They can be learned by maximizing the log-likelihood
\begin{equation} \label{eq1}
   \mathcal{L} = \sum_{i=1}^{N} \sum_{t=t_0}^{T} \log \ell(z_{i,t} \vert \theta(\textbf{h}_{i,t})).
\end{equation} 
Since $h_{i,t}$ is a deterministic function of the input, all quantities needed to compute (\ref{eq1}) are observed, so (\ref{eq1}) can be optimized directly via stochastic gradient descent. In this process, the gradients will be computed with respect to $\Theta$.

For a given dataset, we create multiple training instances for each time series by selecting windows with different starting points from the original time series. 

The total length $T$ and the relative lengths of the conditioning and prediction regions are fixed for all training examples. Selecting these windows ensures that the entire prediction range is always covered by the available baseline data. 
If the start point of the time series is chosen before $t = 1$,
the unobserved targets are set to zero.
In this way, the model can learn the behavior of the new time series given all other available features.
Enriching the data with this windowing procedure ensures that information about absolute time is available to the model only through covariates, not through the relative position of $z_{i,t}$ in the time series.

The detailed training is as follows. 
A network with parameters~$\Theta$ has three inputs: 
the covariates $x_{i,t}$, the target value of the previous time step $z_{i,t-1}$, and the network output of the previous time step $\textbf{h}_{i,t-1}$. The network output 
\begin{equation*}
    \textbf{h}_{i,t}=h(\textbf{h}_{i,t-1},z_{i,t-1},\textbf{x}_{i,t},\Theta)
\end{equation*}is then used to calculate the parameters 
\begin{equation*}
    \theta_{i,t}=\theta(\textbf{h}_{i,t},\Theta)
\end{equation*} of the likelihood $\ell(z|\theta)$, which is used to train the model parameters. 
For prediction, the history of the time series $z_{i,t}$ is fed into the network for $t < t_0$, then a sample is drawn in the prediction domain for $t \geq t_0$ 
and fed back for the next point until the end of the prediction range $t=t_0+T$.  
Repeating this prediction process results in multiple traces that represent the joint predicted distribution. 

\subsection{The DeepCAR Method: Incorporate Change Points}

     
Our \emph{DeepCAR} extends the vanilla DeepAR method by incorporating change points. Change points are common: in our pollution data, for example, there are unpredictable breaks when the machine is being repaired. These change points disrupt the normal behavior of the time series; if tolerated, they can harm predictions because the algorithm misinterprets them. Change points can be identified in various ways. 

As mentioned before, a known method to deal with change points is to remove the data or replace it with linear interpolation methods. However, in our case and also for many real datasets, this approach manipulates the dataset and disrupts the time series behavior. Our DeepCAR method does not change the data but adapts the training process as follows. 

The vanilla DeepAR learns by taking a random batch. A start index is chosen; then a batch with a fixed length is built. For each step, a new start index is chosen.
In our DeepCAR method, we instead successively select windows from the time series, which are chosen to not contain a change point, being able to use the time series as a whole.

In detail, the DeepCAR method searches for the change points before the training starts. When selecting the batches, we define a new function that indicates whether the change point lies in the batch; if the change point lies in the batch, the search is repeated until we find a batch that does not contain a change point. 

\begin{figure}[ht]
\vskip 0.2in
\begin{center}
\centerline{\includegraphics[width=\figurewidth]{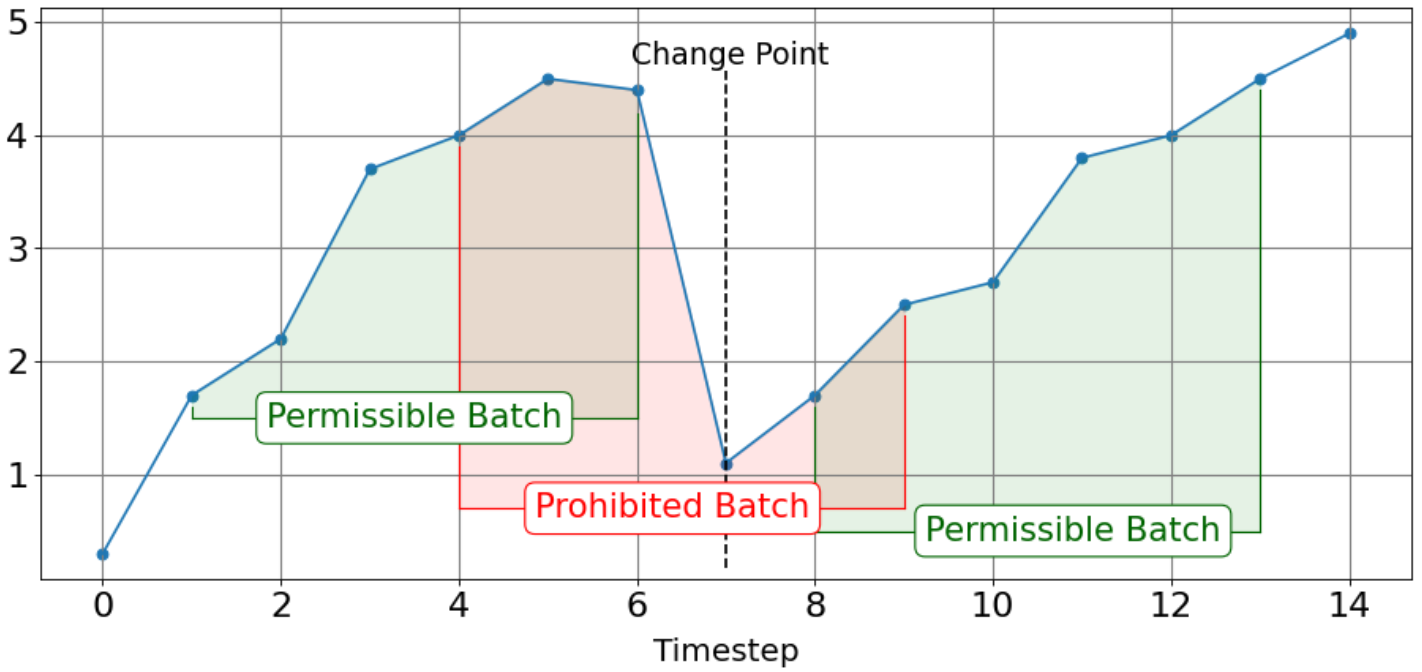}}
\caption{Visualization of the change point method \emph{BatchCP}. It shows an example of a batch selection around a change point. The blue points represent time-series data. The first green area shows a batch created by choosing $t=1$ as the start index, and with a selected batch size of $s=6$, the batch contains the points $t=1$ to $t=6$. The change point is located at $t=7$ and is therefore outside the detected batch. The batch is therefore permissible. The red area next to it shows a prohibited batch. The start index of this batch is $t=4$ and the end index at $t=9$, so the change point is located inside the batch and is not allowed in training. A new batch must be found. The green area to the right indicates another permissible batch, since the change point lies outside of it. }
\label{cpvv}
\end{center}
\vskip -0.2in
\end{figure}

We want to go into more detail about the training process of the DeepCAR.
Just as in vanilla DeepAR, the parameters $\Theta$ of the model are learned by optimizing the log-likelihood (\ref{eq1})
using stochastic gradient descent by computing the gradients with respect to $\Theta$. In the vanilla DeepAR, the batches are selected from the entire time series with a fixed batch size and different start points.
Before we look at the batch generation, we need to mention the change point density and the size of the batches.

It is relevant how large the batch size is selected. It should be chosen depending on the number of change points and the size of the training data set. If there are many change points or if they lie close together due to less training data, the batch size should be chosen accordingly small, so that many batches can be found around the change points and no information gets lost. With small data sets, the batch size should also not be chosen too large, so that enough different batches can be generated. 

In detail, the batch size $s$ is one of the most important control parameters of our method. We have to choose $s$ in such a way that it could cover at most one change point. If $s$ is larger than the distance between two change points $c_i, c_j$, the part of the time series between time index $i$ and $j$  would never be learned. The batches containing this part of the time series would always contain at least one change point and would, therefore, not be allowed. To avoid this, we develop a method that automatically selects the maximum batch size based on the change points. In this method we calculate the distance between all detected change points. We then set $s_{\operatorname{max}}$ to be half of the smallest distance between the change points. We take half of this distance, because we increase the probability of selecting permissible batches between the change points. In addition, the range of the batches is more varied, and we have more diverse training input. This can be seen in Algorithm~\ref{alg:example1}. The number we get for $s_{\operatorname{max}}$ is the maximum batch size that should be selected.
Our experiments have shown that the prediction improves with the choice of a batch size $s\leq s_{\operatorname{max}}$. 


Now, we consider the generation of the batches based on this batch size.
In the DeepCAR method, we pass a batch size $s\leq s_{\operatorname{max}}$ and the detected change points $c_0,..., c_{k-1}$ to the algorithm. Our goal is to find the start and end points of the batches that do not contain any change points. To do this, we select indices that lie between $0$ and $n-s-1$. For each change point, we check whether the index of the respective change point lies between the previously selected start index and end index of the batch. If the change point lies in the batch, we repeat the method and select a new start index, for which the same procedure is performed---see Algorithm~\ref{alg:example}. 

We go through the possible cases in detail: We have found a set of change points 
\begin{equation*}
   C := \{c_0,..., c_{k-1}\}
\end{equation*}and have chosen the batch size $s\leq s_{\operatorname{max}}$ in Algorithm \ref{alg:example1}. We select a random start index $i_{\operatorname{start}}$ which lies in the range $[0,n-s-1]$ of the time series. This gives us a batch with start point $i_{\operatorname{start}}$ and end point 
\begin{equation*}
    i_{\operatorname{end}}:=i_{\operatorname{start}}+s-1\,.
\end{equation*}Assume that one of the change points lies in the batch, thus without loss of generality $\exists  j \in \{0,..., k-1\}$ such that $ c_j \in [i_{\operatorname{start}} , i_{\operatorname{end}}]$. Then again a random start index $i_{\operatorname{start'}}$ is chosen and 
\begin{equation*}
    i_{\operatorname{end'}}:=i_{\operatorname{start'}}+s-1\,.
\end{equation*}If now $\forall j \in \{0,...,k-1\}$ holds 
\begin{equation*}
    c_j \notin [i_{\operatorname{start'}} , i_{\operatorname{end'}}]\,,
\end{equation*} then in the following the batch $[i_{\operatorname{start'}} ; i_{\operatorname{end'}}]$ is a valid batch. 

The choice of the batches is somewhat random. We have ensured that the batches are either completely away from a change point or at the beginning or end of the batch bordering on a change point.
Like the DeepAR, we also use a Gaussian likelihood in the model and apply LSTM layer in addition to the Gaussian layer. In other aspects, the training process follows the DeepAR.

We will  illustrate later in Section~\ref{transformer} that the very same method of including change points also applies to other forecasting methods, such as Transformers or TFTs.
We thus refer to the general method as \emph{\ourmethod}.


\newcommand{\smax}{\ensuremath{s_{\operatorname{max}}}}

\begin{algorithm}[tb]
   \caption{find change points and maximum batch size}
   \label{alg:example1}
\begin{algorithmic}
   \STATE {\bfseries Input:} training data $d$, change point detection method A
   \STATE {\bfseries Output:} maximal batch size $\smax$, change points $c_0,\dots,c_{k-1}$
   \STATE $c_0,\dots,c_{k-1}  \xleftarrow{} A(d)$ ~~~\emph{\footnotesize\# determine change points}
   \FOR{$i,j \in \{0,\dots,k-1\}, i \neq j $}
   \STATE $ \operatorname{diff}_{i,j} \xleftarrow{} \Vert c_i - c_j \Vert_2$ 
   \ENDFOR
   \STATE $\smax \xleftarrow{} \lceil \operatorname{min}_{i,j}\{{\operatorname{diff}_{i,j}}\}/2\rceil$
   \STATE {\bfseries return}  $c_0,\dots,c_{k-1}, \smax$
\end{algorithmic}
\end{algorithm}

\begin{algorithm}[tb]
   \caption{find valid batch}
   \label{alg:example}
\begin{algorithmic}
   \STATE {\bfseries Input:} training data $d$, batch size $s\leq\smax$, \\
   ~~~~~change points $c_0,\dots,c_{k-1}$
   \STATE {\bfseries Output:} start and end points of the batch
   \STATE $n \xleftarrow{} \operatorname{len}(d)$~~~\emph{\footnotesize\# number of samples}
   \REPEAT
   \STATE $i_{\operatorname{start}}  \xleftarrow{}$ random number in $\{0,\dots,n-s-1\}$
   \STATE $\operatorname{valid} \xleftarrow{}\operatorname{True}$
   \FOR{$j \in \{0,\dots,k-1\}$}
   \IF{$c_j \geq i_{\operatorname{start}}$ and $c_j \leq i_{\operatorname{start}}+s-1$}
   \STATE $\operatorname{valid} \xleftarrow{}\operatorname{False}$
   \STATE {\bfseries break for}
   \ENDIF
   \ENDFOR
   \UNTIL{$\operatorname{valid}=\operatorname{True}$}
  \STATE $i_{\operatorname{end}}  \xleftarrow{}i_{\operatorname{start}} +s-1$
   \STATE {\bfseries return} $i_{\operatorname{start}},i_{\operatorname{end}}$

\end{algorithmic}
\end{algorithm}

\subsection{Methods for Change-Point Detection}\label{cpvvc}
Time series can have special points such as anomalies or outliers generated by measurement errors, for example.
These points are usually isolated phenomena and, therefore, relatively straightforward to deal with in terms of prediction:
usually, these points are simply deleted.
But there is also another, more structural type of special points: change points. There are changes in mean, which are the most common change points. There is also change in variance or change in periodicity. A rarer case and also more difficult to detect is change in pattern. Other complex topics are for example changes in multidimensional time series, where changes in correlation are considered. 
Some change points can hardly be recognized visually. Therefore, there are numerous change point detection methods \cite{art2,art8,art11, art12, cpvv,cpvk}. Among others, there is the MOSUM method \citep{deep3}, which is particularly well suited for the detection of the change points of the first type, the change in mean.

We want to talk about the change point detection method MOSUM, that we use in the experiments. For the DeepCAR model, it is especially important to identify the change points as precisely as possible. If this cannot be guaranteed, the DeepCAR method would create batches that contain the actual change points and omit the change points that were found incorrectly. In reality, however, it is often not possible to determine change points manually, because either the data sets are too large, or the determination of the change points is too complex. The idea behind the MOSUM method is the detection of several change points in the mean.
In MOSUM, when the bandwidth $G$ is small, the $\eta$ criterion is used, and when $G$ is large, the $\varepsilon$ criterion is used.
This is because a small value of $G$ is best for highly fluctuating time series to detect small changes, while a large value for $G$ is best for detecting large changes. The disadvantage is that one must specify the $\varepsilon$ criterion for large values of $G$ so that MOSUM is not distracted by neighboring small changes and misses large changes.
\section{Experiments}
This section demonstrates the practical performance of our DeepCAR model.
We consider three real-world data sets:
data from a recycling company (the company already uses our pipeline on a daily basis),
football data,
and treasury data.
We also consider four different scenarios for each data set: (I)~baseline model naïve; 
(II)~ignore all change points, which amounts to vanilla DeepAR, as a reference to show that our model performs as well as  DeepAR in any case;
(III)~include hand-picked change points; and
(IV)~include change points detected by MOSUM to show that our method can largely outperform DeepAR in practice.

Code examples are given in \href{https://github.com/LedererLab/DeepCAR}{https://github.com/LedererLab/DeepCAR}. 

\subsection{Pollution Dataset (Multivariate)}\label{poll}
Our idea is to detect the change points in the time series. We use the change point information when training the model. We consider time series data about oil filter pollution in a shredding machine. Changing the filter changes the data-generating process. We have time series data for one specific machine. The machine data consists of a timestamp with a value for the pollution filter and $10$ additional features, like certain temperatures and pressures, which are dependent on the pollution. These data were prepared in advance to fit the time-series format of the DeepAR model. The goal is to predict the future values for the pollution in order to detect too high values early and to avoid failures and damage to the machine.
We start with a baseline model.

The entire dataset has $10\,200$ rows, each row corresponds to the machine status in a given hour. 
We start with Scenario~(I), the naïve method---see Table~\ref{tablee}. The naïve method is very simple. It takes the current known value and predicts that it will be the same in the future.  
The errors for test and training samples are quite high with this method. 

In the next step, we ignore the change points in the data and run the vanilla DeepAR model \cite{webs}. The batch size is calculated with Algorithm~\ref{alg:example1}. Since the minimum difference is between the change points $c_1 = 1970$ and $c_2 = 2350$, we get the difference $\operatorname{diff}_{1,2} = 380$. Thus, the batch size should be at most 
\begin{equation*}
    \smax=\Bigg\lceil\frac{380}{2}\Bigg\rceil=190\,.
\end{equation*}
We set  $s=30$. 
The loss is the Gaussian likelihood function with parameters~$\mu$ and $\sigma$. The model consists of $3$ layers: a LSTM layer with $4$ units, a dense layer with $3$ units, and a Gaussian layer. ReLU is used as the activation function \cite{DBLP}. 
We do not attempt to optimize the hyperparameters. We split the data set into 60\% training samples, 20\% validation samples, and 20\% test samples. The previous settings will be the same for all methods. 
We calculate the errors as root mean squared error (RMSE). 

For our method in Scenario~(II), 
that is, for vanilla DeepAR,
we get an RMSE of $98.51$ on the training data and $147.83$ on the test data---see the first line in Table~\ref{tablee}.

The change points were trained $265$ times in the training process.
In Scenario~(III), 
we hand-picked the change points of the pollution time series. We know when oil-filter damage occurred to the machine and when the filter was replaced. We picked change points at the following time indices: $700, 1970, 2350, 2730, 3500, 4320, 5050, 6250, 7100$. 

\begin{figure}[ht]
\vskip 0.2in
\begin{center}
\centerline{\includegraphics[width=\figurewidth]{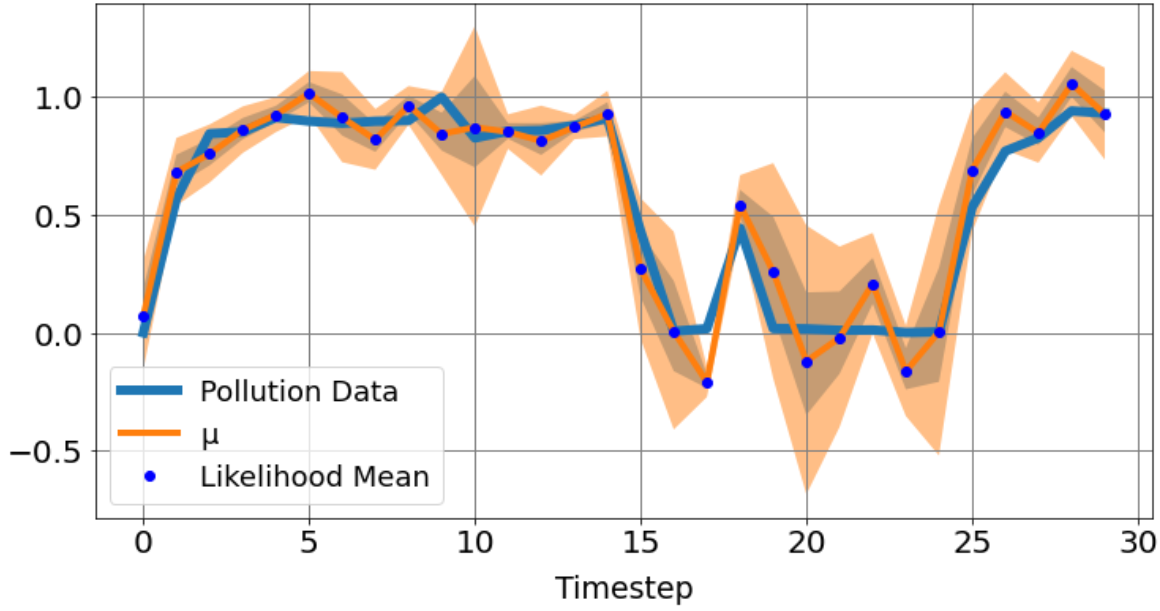}}
\caption{One Batch predicted by the DeepCAR Model. The shown batch has a length of $s=6$, therefore $6$ time steps are shown. The pollution data is blue, the likelihood mean is marked by dark blue data points and $\mu$ corresponds to the orange curve. 
 }
\label{cpp}
\end{center}
\vskip -0.2in
\end{figure}

We pass this list of change points in the training process. When the method that generates the random batches is called, we check whether a change point from our list is in the generated batch. If this is the case, we search for a new valid batch. 
For our method in Scenario~(III), 
we get an RMSE of $81.27$ on the training data and $112.67$ on the test data---see the second line in Table~\ref{tablee}.
Compared to the first line, both the training and test score have improved significantly. We note that both the vanilla DeepAR and DeepCAR methods predict the distribution of the predicted values. This can be seen in Figure \ref{cpp}, where the prediction for one batch from the test data is shown.

\begin{figure}[ht]
\vskip 0.2in
\begin{center}
\centerline{\includegraphics[width=\figurewidth]{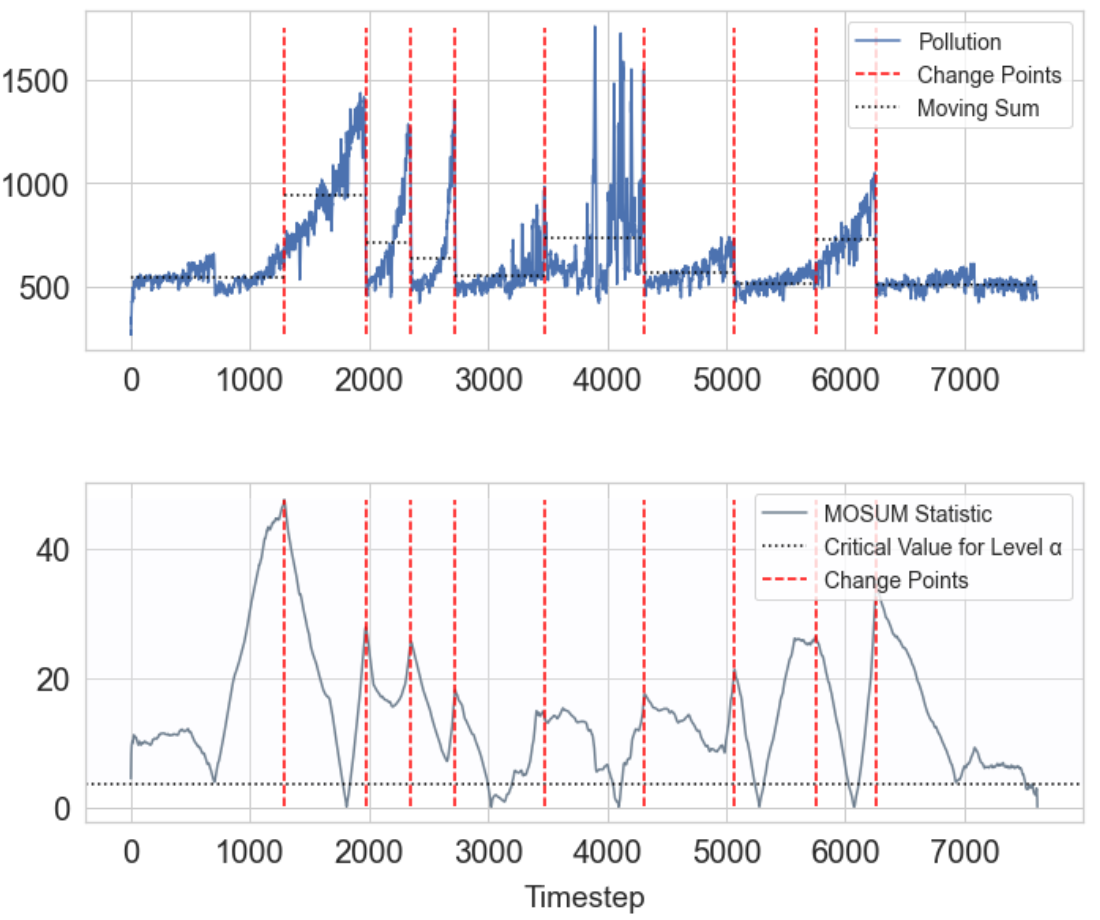}}
\caption{Upper diagram: The pollution time series (blue) in millibar with moving sum (gray) and change points (red) detected with MOSUM. 
Lower diagram: The MOSUM statistic, which identifies the changes in the mean, the threshold level (dotted line), and the corresponding change points (red).
Most of the detected change points equal the actual change points.}
\label{mossum}
\end{center}
\vskip -0.2in
\end{figure}

Now, we include the well-known change-point-detection method MOSUM \cite{deep3}, which consists of procedures for the multiple mean change problem using moving sum statistics---see Section~\ref{cpvvc} .
We choose a bandwidth of $0.2$, and a value for $\eta = 0.1$ and let the MOSUM method detect the change points of the pollution time series. This can be seen in Figure \ref{mossum}.

The resulting list of change points is given to the DeepCAR model and the further procedure is analogous to the previous experiment. For this method (Scenario~(IV)), 
we get a training error of $83.42$ and a test error of $117.75$---see the second line in Table~\ref{tablee}. The error has worsened slightly, which can be explained by the fact that the MOSUM change point detection is not 100\% correct but also forgot some points. The error is still smaller than with the DeepAR without taking the change points into account.

All results of the experiments can be seen in Table \ref{tablee}. The model is clearly performing better than the vanilla deepAR on the same dataset. Our DeepCAR method (Scenario~(III)) is more than three times better than the naïve baseline model (Scenario (I)) and improves the vanilla DeepAR (Scenario~(II)) by 23,78 \%. Even with automatic change point detection (Scenario~(IV)), which could be further optimized, we are  $20,35$ \% better than the vanilla DeepAR. 
\begin{table}[t]
\caption{Prediction error of the different methods on the pollution data. 
Reported are test and training RMSE of (I) naïve, (II)~DeepAR and our DeepCAR equipped with (III)~manual and (IV)~MOSUM change-point detection.}
\label{sample-table}
\vskip 0.15in
\begin{center}
\begin{small}
\begin{sc}
\begin{tabular}{lcc}
\toprule
Scenario & Train RMSE & Test RMSE  \\
\midrule
~~\,(I)~~\,\!\,\,\!Baseline naïve &  198.55 &	382.08\\
\,\,\,(II)~~\,\!\,\!DeepAR-No CPD & ~~98.51& 147.83\\
~(III)~\,DeepCAR-Manual& ~~81.27  & 112.67\\
~(IV) \,DeepCAR-MOSUM& ~~83.42  & 117.75  \\
\bottomrule
\end{tabular}

\end{sc}
\end{small}

\end{center}

\vskip -0.1in
\label{tablee}
\end{table}

\subsection{Football Dataset (Univariate)}

\begin{figure}[ht]
\vskip 0.2in
\begin{center}
\centerline{\includegraphics[width=\figurewidth]{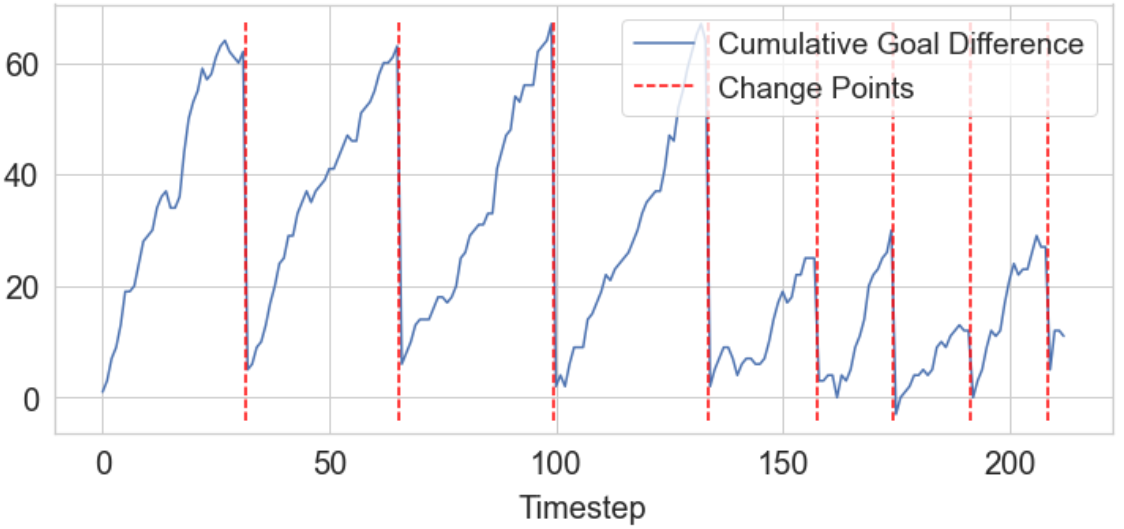}}
\caption{Football data (blue) with manual change points (red) located at time indices 31, 65, 99, 133, 157, 174, 191, 208.}
\label{soccer11}
\end{center}
\vskip -0.2in
\end{figure}

Next, we consider a \href{https://www.football-data.co.uk/germanym.php}{football dataset} \cite{1}. The dataset consists of a date---the match day---and the difference in goals scored and received by a team. After each season, the statistics are reset, resulting in the change points in the dataset. The dataset is univariate since there are no features. The dataset with change points can be seen in Figure \ref{soccer11}. The change points are at time indices 31, 65, 99, 133, 157, 174, 191, 208.
We again use the naïve method (Scenario I) as a baseline model---see Table~\ref{tablee2}.

As already mentioned, it is relevant how large the batch size is selected. In the case of the football dataset, we only have a small amount of training data, so the batch size should be correspondingly smaller. 
We again calculate the batch size with Algorithm~\ref{alg:example1}. The minimum difference is between the change points $c_4 = 157$ and $c_5 = 174$, we get the difference $\operatorname{diff}_{4,5} = 17$. Thus, the batch size should be at most 
\begin{equation*}
    s_{\operatorname{max}}=\Bigg\lceil \frac{17}{2}\Bigg\rceil = 9\,.
\end{equation*} We apply this batch size.
 
We now start with our method in Scenario~(II), 
that is, for vanilla DeepAR,
we get a RMSE of $15.41$ on the training data and $51.78$ on the test data---see the first line in Table~\ref{tablee2}.

Next, we use our method in Scenario~(III) with the change points we hand-picked manually before. The change points are passed into the training. The procedure is the same as in the pollution experiment, except that this time, we have a univariate time series.
We get an RMSE of $8.20$ on the training data and $14.89$ on the test data---see the second line in Table~\ref{tablee2}.
Compared to the first Scenario, both the training and test score have improved significantly.

\begin{figure}[ht]
\vskip 0.2in
\begin{center}
\centerline{\includegraphics[width=\figurewidth]{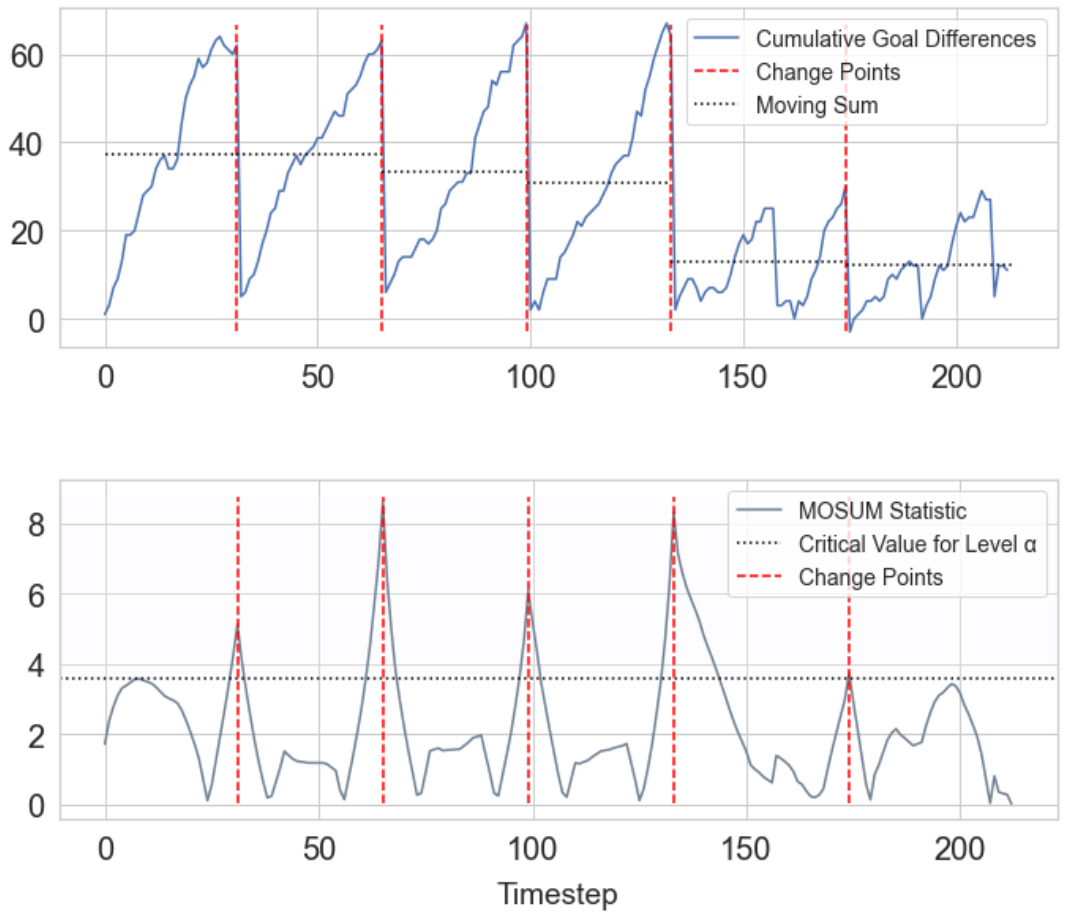}}
\caption{Upper diagram: The football data (blue) with
change points (red) detected with MOSUM, and the corresponding moving sum (grey). Lower diagram: The MOSUM statistic, which identifies the changes in the mean, and the corresponded change points in red. The defined threshold value is also plotted in grey.}
\label{soccer2m}
\end{center}
\vskip -0.2in
\end{figure}

Now we again use MOSUM to detect the change points of the football dataset---see Figure \ref{soccer2m}. 

The detection is very good: most manually-set change points are detected. We pass this list of change points into the training of the DeepCAR.
Thus for Scenario~(IV) we get a training error of $10.44$ and a test error of $18.63$. 

We again compare all results of the football data experiments in Table \ref{tablee2}. We again use the test error for a comparison. The results with our DeepCAR method are again significantly better than with naïve and with vanilla DeepAR. The test error with DeepCAR  (Scenario~(III)) is 41.88 \% better than the vanilla DeepAR. With MOSUM (Scenario~(IV)) we are 27.09 \% better than the vanilla DeepAR. 



\begin{table}[t]
\caption{Prediction error of the different methods on the football data. 
Reported are test and training RMSE of (I) naïve, (II)~DeepAR and our DeepCAR equipped with (III)~manual and (IV)~MOSUM change-point detection.}

\vskip 0.15in
\begin{center}
\begin{small}
\begin{sc}
\begin{tabular}{lcc}

\toprule
Scenario & Train RMSE & Test RMSE  \\
\midrule
~~\,(I)~~\,\!\,\,\!Baseline naïve & $25.29$& $51.78$ \\
\,\,\,(II)~~\,\!\,\!DeepAR-No CPD & $15.41$& $25.62$\\
~(III)~\,DeepCAR-Manual & $~~8.20$&  $14.89$\\
~(IV) \,DeepCAR-MOSUM & $10.44$& $18.63$   \\

\bottomrule
\end{tabular}

\end{sc}
\end{small}
\end{center}

\vskip -0.1in
\label{tablee2}
\end{table}

\subsection{Treasury Rate (Univariate)}
We now consider univariate data of the Federal Reserve Board~\cite{23}.
The samples are daily averages of yields of several treasury securities, all adjusted to the equivalent of a one-year maturity.

With our baseline modell (I) we get an RMSE of $3.98$ for training samples and $8.17$ for test samples. 

\begin{figure}[ht]
\vskip 0.2in
\begin{center}
\centerline{\includegraphics[width=\figurewidth]{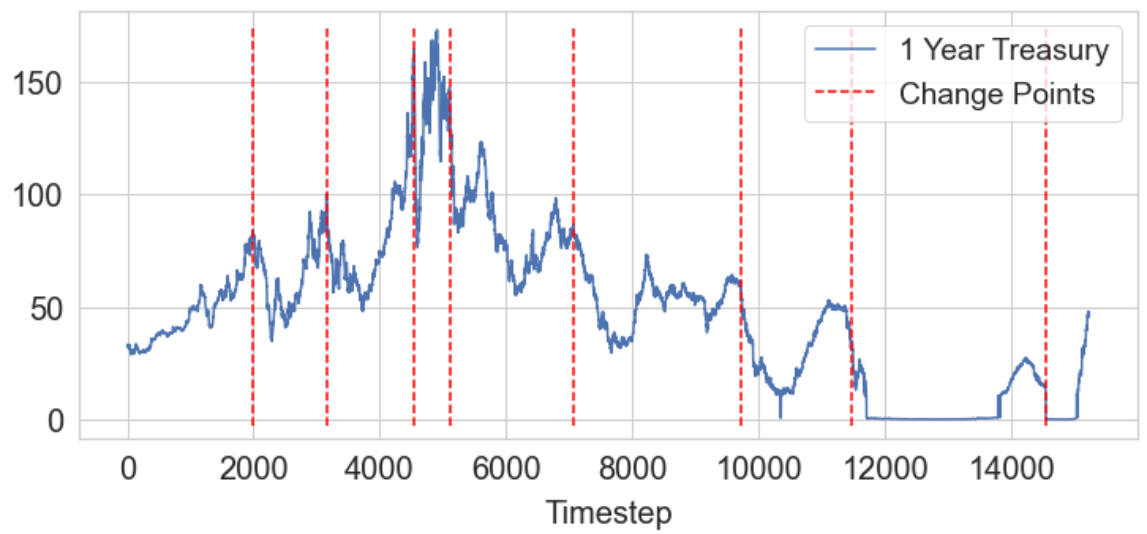}}
\caption{Treasury dataset (blue) with manual change points (red) located at time indices 1992, 3155, 4544, 5105, 7065, 9710, 11480, 14543.}
\label{steddo}
\end{center}
\vskip -0.2in
\end{figure}

The change points are are not immediately recognizable here. 
But we know that recessions have occurred at time points 1992, 3155, 4544, 5105, 7065, 9710, 11480, 14543---see Figure~\ref{steddo}. 
MOSUM recognizes these change points relatively accurately again (for detailed MOSUM results of the treasury dataset see Appendix~\ref{appendix}). 
Algorithm~\ref{alg:example1} yields the maximal batch size 
\begin{equation*}
    \smax=\biggl\lceil \frac{4544-3155}{2}\biggr\rceil=\biggl\lceil \frac{1389}{2}\biggr\rceil = 695\,.
\end{equation*}
We choose $s = 50$. 

The results are summarized in Table~\ref{tableee}.
Once more, our DeepCAR outmatches DeepAR:
the test errors of DeepCAR  are about $9\,\%$ (Scenario~(III)) and $5\,\%$ (Scenario~(IV) better than for DeepAR and also much better than for naïve.



\begin{table}[t]
\caption{Prediction error of the different methods on the Treasury data. 
Reported are test and training RMSE of (I) naïve, (II)~DeepAR, and our DeepCAR equipped with (III)~manual and (IV)~MOSUM change-point detection.}
\vskip 0.15in
\begin{center}
\begin{small}
\begin{sc}
\begin{tabular}{lcc}
\toprule
Scenario & Train RMSE & Test RMSE \\
\midrule

~~\,(I)~~\,\!\,\,\!Baseline naïve &  $3.98$ &$8.17$ \\
\,\,\,(II)~~\,\!\,\!DeepAR-No CPD &$4.13$ & $5.93$ \\
~(III)~\,DeepCAR-Manual & $3.40$  & $5.37$\\
~(IV) \,DeepCAR-MOSUM &  $3.69$ & $5.61$  \\

\bottomrule
\end{tabular}
\end{sc}
\end{small}
\end{center}

\vskip -0.1in
\label{tableee}
\end{table}

\subsection{Synthetic data}
\begin{figure}[ht]
\vskip 0.2in
\begin{center}
\centerline{\includegraphics[width=\figurewidth]{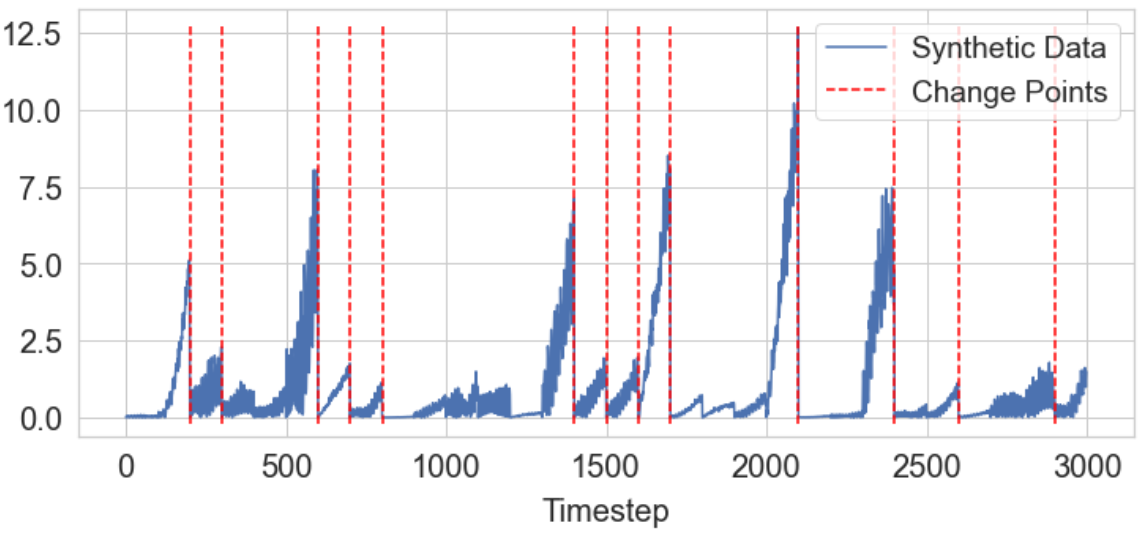}}
\caption{Synthetic data (blue) with manual change points (red) located at time indices 200, 300, 600, 700, 800, 1400, 1500, 1600, 1700, 2100, 2400, 2600, 2900.}
\label{synthetic}
\end{center}
\vskip -0.2in
\end{figure}

We substantiate the empirical results further by considering synthetic data,
where we know and control the data-generating process.
The data comprises $3000$~samples with $13$~change points---see Figure \ref{synthetic}. 
The naïve method (I) yields $3.57$ RMSE for the training samples and $9.33$ for the test samples. 

Algorithm~\ref{alg:example1} allows for a batch size of $s=50$.
The results for Scenarios~(I), (II), (III), and (IV) are in Table~\ref{tableee5}. 
DeepCAR, both with manually and automatically selected change points, outmatches vanilla DeepAR and naïve once more.

\begin{table}[t]
\caption{Prediction error of the different methods on the synthetic data. 
Reported are test and training RMSE of (I) naïve, (II)~DeepAR and our DeepCAR equipped with (III)~manual and (IV)~MOSUM change-point detection.}
\vskip 0.15in
\begin{center}
\begin{small}
\begin{sc}
\begin{tabular}{lcc}
\toprule
Scenario & Train RMSE & Test RMSE \\
\midrule
~~\,(I)~~\,\!\,\,\!Baseline naïve & $3.57$ &$9.33$\\
\,\,\,(II)~~\,\!\,\!DeepAR-No CPD & $2.81$& $5.77 $ \\
~(III)~\,DeepCAR-Manual & $2.42 $  & $ 5.40 $\\
~(IV) \,DeepCAR-MOSUM & $2.51 $  & $ 5.62 $  \\
\bottomrule
\end{tabular}
\end{sc}
\end{small}
\end{center}

\vskip -0.1in
\label{tableee5}
\end{table}
\section{Extensions Beyond DeepAR} \label{transformer}
So far, we have extended DeepAR.
But importantly,
our concepts apply to batch-trained methods much more generally.
Examples are the attention-based Transformer or Temporal Fusion Transformer (TFT) \citep{Natt,tft}.
To illustrate this,
we implement Transformer via Gluon Time Series (GluonTS) \cite{glu}.
The batches are generated according to Algorithm~\ref{alg:example}.

We test the pipeline on the pollution data of Section~\ref{poll}.
We consider the following three scenarios: (A)~ignore all change points;
(B)~include hand-picked change points (the same ones as in Section~\ref{poll}); and
(C)~include change points detected by MOSUM (the same ones as in Section~\ref{poll}).
The results are summarized in Table~\ref{table6}.
We observe that the Transformer model generally seems to work better than the DeepAR model in this example---compare to Table~\ref{tablee}---but this is not necessarily the case in other examples. 
Important for us is that we can  improve the Transformer noticeably by our \ourmethod\ method for including change points.

\begin{table}[t]
\caption{Prediction errors on the pollution data. 
Reported are test and training RMSE of (A) Transformer without considering change points and Transformer equipped with (B)~manual, and (C)~MOSUM change-point detection.}
\vskip 0.15in
\begin{center}
\begin{small}
\begin{sc}
\begin{tabular}{lcc}
\toprule
Scenario & \hspace{-3mm}Train RMSE & \hspace{-1mm}Test RMSE \\
\midrule
\hspace{-3mm}~(A)~Transformer-No CPD & \hspace{-3mm}$75.26$ &\hspace{-1mm}$96.23$\\
\hspace{-3mm}~(B)~Transformer-Manual & \hspace{-3mm}$67.85$& \hspace{-1mm}$91.02$ \\
\hspace{-3mm}~(C)~Transformer-MOSUM & \hspace{-3mm}$70.22$  & \hspace{-1mm}$93.12$\\
\bottomrule
\end{tabular}
\end{sc}
\end{small}
\end{center}

\vskip -0.1in
\label{table6}
\end{table}

\section{Conclusion}
Our generalization of DeepAR, called DeepCAR, 
improves on the vanilla version substantially when there are change points and equals that version if no change points are detected.
Thus, DeepCAR can generally be used as a drop-in replacement for DeepAR.

But equally importantly, our work identifies the batch size as a way to account for change points in time-series analyses more generally.
As an illustration, we have shown that our general method to account for change points, called DeepCAR,
can improve Transformers as well.

\section{Acknowledgments}

We sincerely thank Kata Vuk and David Salinas for their helpful input.

\bibliography{References}
\bibliographystyle{icml2021}

\appendix
\onecolumn
\section{Appendix} \label{appendix}

\begin{figure}[ht]
\vskip 0.2in
\begin{center}
\centerline{\includegraphics[width=\figurewidth]{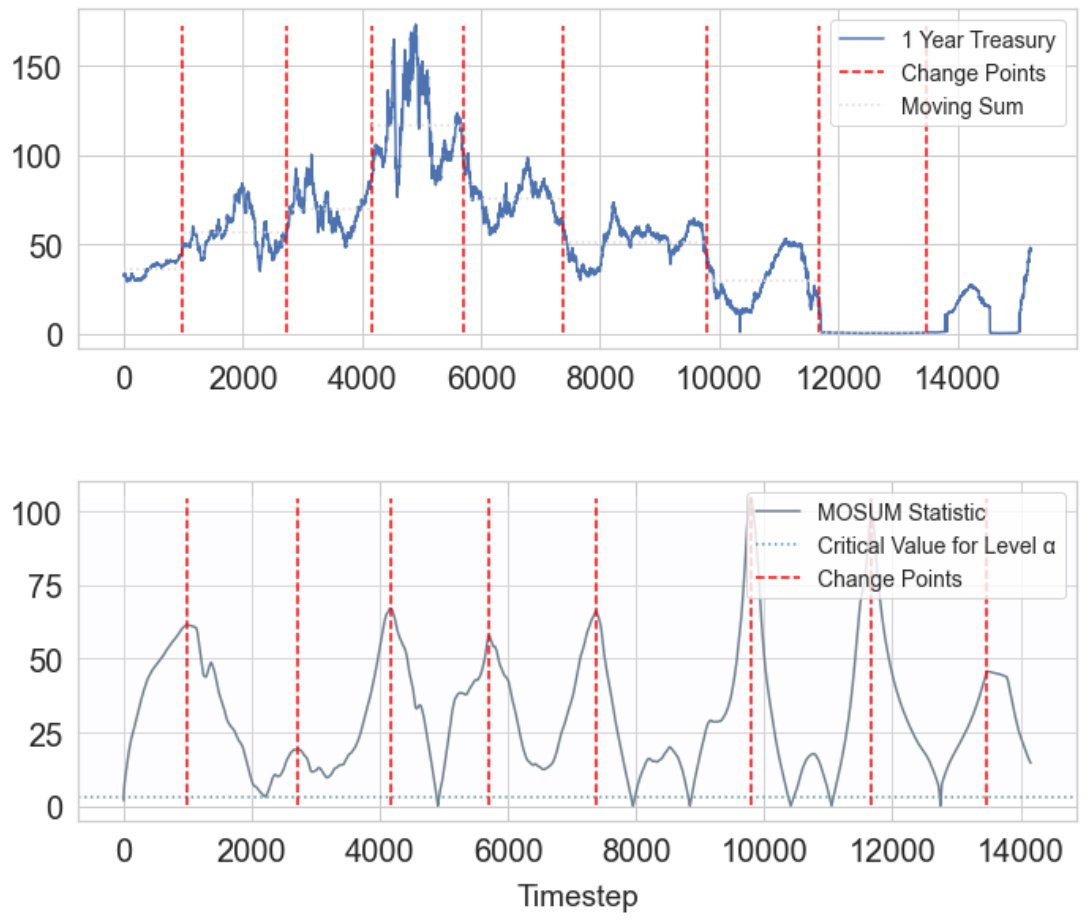}}
\caption{Upper diagram: The one-year-treasury time series (blue) with
change points (red) detected with MOSUM, and the corresponding moving sum (grey). Lower diagram: The MOSUM statistic, which identifies the changes in the mean, and correspondingly the detected change points in red. The defined threshold value is plotted in grey.}
\label{stedrt}
\end{center}
\vskip -0.2in
\end{figure}

Figure~\ref{stedrt} shows the treasury data with change points detected by MOSUM. 

\end{document}